# The Optimal ANN Model for Predicting Bearing Capacity of Shallow Foundations Trained on Scarce Data

*Marta Bagińska[1] and Piotr E. Srokosz[2]*

**ABSTRACT:**

**This study is focused on determining the potential of using deep neural networks (DNNs) to predict the ultimate bearing capacity of shallow foundation in situations when the experimental data which may be used to train networks is scarce. Two experiments involving testing over 17000 networks were conducted. The first experiment was aimed at comparing the accuracy of shallow neural networks and DNNs predictions. It shows that when the experimental dataset used for preparing models is small then DNNs have a significant advantage over shallow networks. The second experiment was conducted to compare the performance of DNNs consisting of different number of neurons and layers. Obtained results indicate that the optimal number of layers varies between 5 to 7. Networks with less and – surprisingly – more layers obtain lower accuracy. Moreover, the number of neurons in DNN has a lower impact on the prediction accuracy than the number of DNN's layers. DNNs perform very well, even when trained with only 6 samples. Basing on the results it seems that when predicting the ultimate bearing capacity with ANN models obtaining small but high-quality experimental training datasets instead of large training datasets affected by a higher error is an advisable approach.**

*Keywords: ultimate bearing capacity, shallow foundations, heuristic algorithms, artificial neural networks, deep learning*

[1] Master Student, The Faculty of Geodesy, Geospatial and Civil Engineering,
University of Warmia and Mazury,
Prawocheńskiego 15, 10-720 Olsztyn, Poland
mail to: marta.baginska.1@student.uwm.edu.pl

[2] Professor, The Faculty of Geodesy, Geospatial and Civil Engineering,
University of Warmia and Mazury,
Prawocheńskiego 15, 10-720 Olsztyn, Poland
mail to: psrok@uwm.edu.pl

## 1. Introduction

Artificial neural networks (ANNs) have attracted wide interest in many practical geotechnical problems. Among them, the problem of estimating the load bearing capacity of shallow foundations alongside determining the load bearing capacity of pile foundations. This problem is an important issue for example in case of transportation constructions if the bearing capacity of the underlying soils is not sufficient to support the construction. In such case one needs effective techniques for estimating the dimensions of foundations will which not only ensure the safety of the object being constructed but also generate low construction and operation costs.

The implementation of the ANN models in geotechnical problems has drawn much attention. MURAT ORNEK et al. (2012) successfully used ANNs and the multi-linear regression model (MLR) to predict the bearing capacity of circular shallow footings supported by layers of compacted granular fill over natural clay soil. KUO et al. (2009) developed a bearing capacity prediction method based on multiple regression methods and multilayer perceptrons (MLPs, ANN). Obtained equations were compared with traditional methods. It was shown that ANN outperforms the existing methods.

The issues of load bearing capacity of shallow foundations can be found in many studies on the use of genetic algorithms and genetic programming. The study presented by SHAHNAZARI & TUTUNCHIAN (2012) concentrated on using multigene Genetic Programming (GP) soft computing technique to predict the ultimate bearing capacity of shallow foundations on cohesionless soils. The developed GP-based formula was calibrated and validated using an experimental database, which was obtained from full-scale foundations load tests and from small-scale laboratory footing load tests. CHAN-PING PAN et al. (2013) used weighted genetic programming (WGP) and soft computing polynomials (SCP) to provide accurate prediction and formulas for the ultimate bearing capacity. SHERVIN TAJERI et al. (2015) and ALAVI & SADROSSADAT (2016) proposed a novel formulation for the ultimate bearing capacity of shallow foundations resting on/in rock masses, using a linear genetic programming. The derived models reached a notably better prediction performance than the traditional equations.

Researchers are more and more willing to use ANFIS models to solve the problems of bearing capacity of foundations. PROVENZANO et al. (2004) showed the possibilities of ANFIS application to determine load–settlement curves diagrams for predicting footing foundation behavior subjected to vertical centered and eccentric loads. The system was trained by the results of a series of small scale load tests both in normal and gradually increased gravity conditions. This path of searching for effective models is being intensively developed (see PADMINI et al. (2008), ADEM KALINLI et al. (2011), MOAYEDI & HAYATI (2018)). The results clearly indicate, that the developed AI soft computing models have excellent agreement with *in-situ* observations and perform better than the theoretical methods.

Artificial neural networks were also extensively tested in pile foundations designing problems. BYUNG TAK KIM et al. (2001) applied ANN to the prediction of lateral behavior of single and group piles. BAZIAR et al. (2015) performed ANN analyses to obtain accurate model for the pile settlement prediction. A similar issue was considered by POOYA NEJAD & JAKSA (2017) - the application of ANN generated a series of charts for predicting pile behavior which can be useful in pile design problems. MILAD et al. (2015) used ANN to predict pile capacity with respect to the flap number (number of hammer strikes), soil parameters, and pile geometry.

Further analyses performed with aid of genetic programming and linear regression allowed to derive equations describing pile bearing capacity. As a final result, they proposed a new method to predict the compressive bearing capacity of driven piles. BASHAR TARAWNEH (2013) developed an ANN model for predicting time dependent capacity of piles (known as pile setup) and demonstrated that the new model significantly outperforms the examined empirical formulas. The issue of pile setup was continued by BASHAR TARAWNEH & RANA IMAM (2014), who conducted analyzes based on Multiple Linear Regression (MLR) and ANN for three different pile types (pipe, concrete, and H-pile) using dynamic load tests results. ISMAIL & JENG (2011) developed a high-order neural network (HON) to simulate the pile load–settlement curve.

All applications of ANN in pile designing problems showed the superiority of the proposed models over standard methods. But ANNs were widely used also in other geotechnical problems, like prediction of the uplift capacity of suction foundations - offshore caissons (RAHMAN et al., 2001), estimation of undrained cohesion intercept of soil in relation with soil physical properties (MOLLAHASANI et al., 2011), estimation of subgrade resilient modulus in correlation with the stress state and physical properties of soils (SUNG-HEE KIM et al., 2014), correlation of intensity of permeation grouting with the physical parameters and the water-to-cement ratio (KUO-WEI LIAO et al., 2011), prediction of soil pore-water pressure responses to rainfall in terms of slope stability problem (MUSTAFA et al., 2012), correlation between soils initial parameters and the strain energy required to trigger liquefaction in sands and silty sands BAZIAR & JAFARIAN (2007), ANFIS models for predicting the bearing capacity of stone columns MANITA DAS & ASHIM KANTI DEY (2017).

It should be stressed that the studies described above showed that evolutionary algorithms (EA) like learning classifier systems (LCS), artificial neural networks (ANN) or adaptive neural-fuzzy inference systems (ANFIS) are very effective in predicting the ultimate bearing capacity of shallow foundations. But there is one important issue that in many cases may prevent the use of EAs in practice. All the mentioned methods were build using comprehensive sets of experimental data which is used to train algorithms. Such sets may be obtained in two ways:

1. By conducting laboratory experiments. In such case it is relatively easy to obtain large experimental datasets. But algorithms trained on these datasets may not perform well in predicting bearing capacity in real-life applications. For example, the change of scale may have impact on the error of predictions generated by the model (ABU-FARSAK et al., 2008).
2. By conducting many large-scale experiments, which may prove to be economically ineffective.

The study described in the present paper is focused on finding the optimal model of ANNs for training on a very *small* dataset. In more detail deep artificial neural networks (DNNs) with different number of layers are trained on experimental datasets representing the ultimate bearing capacities of foundations. Then the accuracy of DNNs trained on small and large datasets is compared. Next the optimal architecture for networks trained on small datasets is studied.

The results of this study may be used in designing ANN models (and hybrid models incorporating ANNs like ANFIS models) for predicting ultimate bearing capacity in situations when the experimental training data is scarce.

**Remark.** We would like to mention that this study is motivated by the following observation. Science and technology development has always been inspired by the conceptual solutions observed in nature. One of the most interesting examples of such phenomenon are optimization techniques such as evolutionary algorithms (EA), learning classifier systems (LCS), artificial neural networks (ANN) or adaptive neural-fuzzy inference systems (ANFIS) used in machine learning. In general, all techniques that lead to the increase of the quality of machine-performed task with the increase of the gained experience can be called artificial intelligence (AI) (MITCHELL, 1997). The need of improving the comfort of life on the one hand and generating time and financial savings on the other induces the necessity of using artificial intelligence in many areas, including engineering. This paper focuses on the important problem encountered in geotechnical engineering: automating the estimation of the load bearing capacity of a shallow foundation using ANNs.

## 2. Deep structured learning

An artificial neural network (ANN) is a brain-inspired computing system consisting of interconnected processing units (neurons), designed to model input-output dependencies in a training process and predict values of approximated multivariable functions.

The network learns (increases prediction accuracy) by analyzing multiple datasets and adjusting connection weights. The accuracy of the prediction is often corelated with the amount of data used for training.

*Definition of ANN*
In this paper an ANN is defined as a function:
$$ANN: X \to Y,$$
where
$X \subset \mathbb{R}^m$ is a space of inputs,
$Y \subset \mathbb{R}^k$ is a space of outputs.

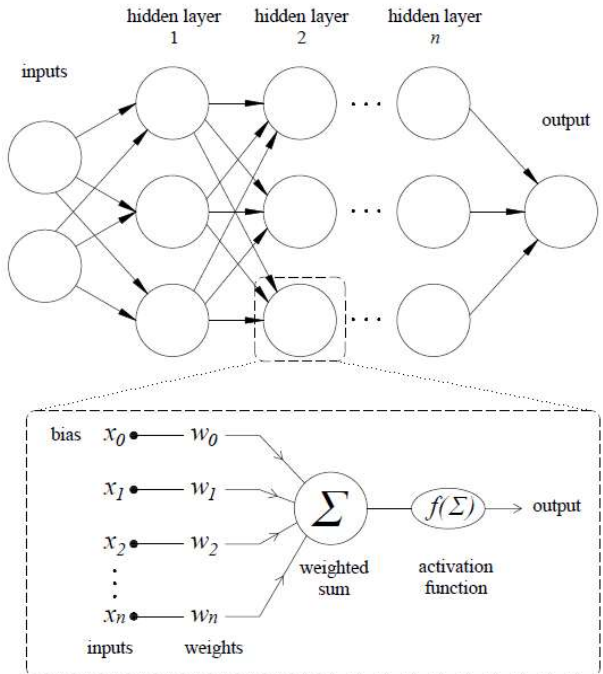

**Figure 1.** Deep artificial feedforward neural network diagram

A model of a deep neural network is shown on **Figure 1.** In this case $ANN: \mathbb{R}^2 \to \mathbb{R}$. Input vectors are processed by neurons within the first layer. Each coordinate of the input vector is multiplied by a corresponding weight. Neuron sums up those products and calculates a value of activation function. The data is forwarded to the next layers until the output is obtained.

An activation function is a real function, which determines the value returned by the neuron. Most frequently used functions are logistic function, hyperbolic functions (tanh), inverse trigonometric functions (arctan), softsign.

There are two primary types of ANN's architecture: feedforward neural networks (described above) and recurrent neural networks. In recurrent ANNs connections between neurons can form loops, which allows the information to be processed multiple times by the same neuron.

The most common architecture of feedforward neural network is multilayer perceptron (MLP). It consists of at least three layers: an input layer, an output layer, and a hidden layer. An MLP with multiple hidden layers is called the deep neural network (DNN). Additional layers and therefore connections allow to model rare dependencies in training data more precisely using fewer neurons (BARAL et al., 2018). However, the learning process of a DNN may lead to overfitting and performance decline as a result (COGSWELL, 2016).

The universal approximation theorem (CYBENKO, 1989, HORNIK, 1991), which plays a central role in theory of artificial neural networks states that a single hidden layer MLP is sufficient to estimate any compactly supported continuous real function with a given accuracy. However, as practical research shows (HE et al., 2016), in many cases DNNs' predictions are more accurate than the ones obtained by shallow networks.

In the training process an ANN modifies weights according to a gradient of an error function, so that the error is minimalized. There are many different algorithms that can be used for training. The algorithms may differ in terms of performance, depending on a specificity of a given problem (SCHMIDHUBER, 2015).

In this study the Bayesian regularization backpropagation algorithm was used to train the ANNs. Pre-research simulations showed a prominent predominance of this algorithm over other tested training functions (Levenberg-Marquardt optimization, gradient descent, quasi-Newton backpropagation). The learning algorithm of Bayesian regularization manipulates the weights corresponding to neuron connections and biases according to Levenberg-Marquardt optimization method to minimize a linear combination of squared errors and weights.

## 3. Description of experiments and the architecture of DNNs used

The ANN models were programmed in MATLAB environment using Neural Network Toolbox 10 (MathWorks 2017).

The first experiment was aimed to compare the accuracy of the predictions of shallow and deep neural networks trained on *small* and *large* training sets. To eliminate the impact of training parameters on the performance of shallow neural networks different combinations of number of neurons, activation functions and training algorithms were used for creating shallow MLPs.

The architecture of created MLPs is presented in **Table 1.** Twenty DNNs of each architecture were generated to compare their performance with 2000 generated shallow neural networks.

Table 1. The architecture of MLPs used in the first experiment.

|  | number of hidden layers | number of neurons in *hidden* layers | number of trained networks (for a given training set) |
|---|---|---|---|
|  | 1 | 90 to 200 | 2 000 |
| deep neural networks | 2 | 45-45 | 20 |
|  | 3 | 30-30-30 | 20 |
|  | 4 | 22-24-22-22 | 20 |
|  | 5 | 18-18-18-18-18 | 20 |
|  | 6 | 15-15-15-15-15-15 | 20 |
|  | 7 | 13-13-13-13-13-13-12 | 20 |

Optimal choice of training parameters: activation functions, the learning rate and the training algorithm is crucial for the effective training of a network. For DNNs the log-sigmoid transfer function was used as an activation function. In pretesting this function proved to be the best (in terms of the accuracy of the prediction) among available in MATLAB neuron activation functions: tan-sigmoid, positive linear transfer function.

The learning rate was set to be $\eta = 0.5$, as the smaller values of $\eta$ caused the training algorithm finding only local minima of the error functions (what resulted in large errors in predictions).

All the DNNs were trained using the Bayesian regularization algorithm. The shallow networks were trained using the following algorithms: Levenberg-Marquardt optimization, gradient descent, quasi-Newton backpropagation and Bayesian regularization.

To minimize the overfitting effect in DNNs training introducing a learning-stopping criterion is necessary. In this study, the training process was stopped after 100 epochs or when the validation error has increased 10 times straight.

The second experiment was aimed to compere DNNs (trained on small set consisting of only 6 samples) with different number of layers and neurons. The architecture of generated DNNs is presented in **Table 2**. For each number of neurons (90,120,150) 2000 shallow MLPs were trained. Also, for each number of neurons 20 DNNs were trained with 2,3,4,5,6,7,8,9,10 and 15 hidden layers.

Table 2. The architecture of DNNs used in the second experiment.

|  | number of neurons | 90 | 120 | 150 |
|---|---|---|---|---|
|  |  | DNN's architecture | | |
| number of hidden layers | 1 | 90 | 120 | 150 |
|  | 2 | 2x45 | 2x60 | 2x75 |
|  | 3 | 3x30 | 3x40 | 3x50 |
|  | 4 | 2x23-2x22 | 4x30 | 2x38-2x37 |
|  | 5 | 5x18 | 5x24 | 5x30 |
|  | 6 | 6x15 | 6x20 | 6x25 |
|  | 7 | 6x13-12 | 6x17-18 | 3x22-4x21 |
|  | 8 | 6x11-2x12 | 8x15 | 6x19-2x18 |
|  | 9 | 9x10 | 6x13-3x14 | 6x17-3x16 |
|  | 10 | 10x9 | 10x12 | 10x15 |
|  | 15 | 15x6 | 15x8 | 15x10 |

## 4. Data

In many studies (for example PADMINI et al., 2008; KALINI et al., 2011), where MLPs' effectiveness of predicting the ultimate bearing capacity of shallow foundation was tested, a wide range of data was used to train artificial neural networks. This dataset contained results obtained in experiments conducted in different laboratories (MUHS et al., 1969, WEISS, 1970, MUHS & WEISS, 1971, 1973, BRIAUD & GIBBENS, 1999, GANDHI, 2003). It should be stressed once again that the aim of this study was to find types of MLPs that can be trained most effectively on small data samples. The usefulness of neural networks (trained on relatively large experimental datasets) in predicting bearing capacity is, as it was mentioned in Introduction, already known and well documented.

Therefore, to eliminate non-defined quantitative and qualitative differences between the foundation models and the measuring systems in different experiments the dataset used in this study was limited to the series of experiments conducted by GANDHI (2003).

Table 3. The data used for the ANN model development (GANDHI, 2003)
(B,D,L-width, depth and length of the foundation; γ,Φ-unit weight and internal friction angle of soil, $q_u$ - bearing capacity)

| B (m) | D (m) | L/B (-) | γ (kN/m³) | Φ (°) | $q_u$ (kPa) | |
|---|---|---|---|---|---|---|
| 0.0585 | 0.029 | 5.95 | 15.7 | 34 | 58.5 | |
| 0.0585 | 0.058 | 5.95 | 17.1 | 42.5 | 211 | |
| 0.094 | 0.047 | 6 | 16.5 | 39.5 | 155.8 | |
| 0.094 | 0.094 | 6 | 17.1 | 42.5 | 279.6 | |
| 0.152 | 0.075 | 5.95 | 15.7 | 34 | 98.2 | |
| 0.152 | 0.15 | 5.95 | 17.1 | 42.5 | 400.6 | A |
| 0.094 | 0.047 | 1 | 16.1 | 37 | 98.8 | |
| 0.094 | 0.094 | 1 | 17.1 | 42.5 | 295.6 | |
| 0.152 | 0.15 | 1 | 16.5 | 39.5 | 264.5 | |
| 0.094 | 0.047 | 1 | 15.7 | 34 | 67.7 | |
| 0.152 | 0.075 | 5.95 | 16.5 | 39.5 | 211.2 | |
| 0.0585 | 0.058 | 5.95 | 16.5 | 39.5 | 142.9 | |
| 0.152 | 0.15 | 1 | 17.1 | 42.5 | 423.6 | B |
| 0.152 | 0.075 | 1 | 15.7 | 34 | 91.2 | |
| 0.094 | 0.047 | 6 | 16.1 | 37 | 104.8 | |
| 0.0585 | 0.029 | 5.95 | 16.1 | 37 | 82.5 | |
| 0.094 | 0.047 | 6 | 15.7 | 34 | 74.7 | |
| 0.152 | 0.15 | 5.95 | 16.8 | 41.5 | 342.5 | |
| 0.094 | 0.047 | 6 | 16.8 | 41.5 | 206.8 | |
| 0.152 | 0.075 | 1 | 16.1 | 37 | 135.2 | C |
| 0.094 | 0.047 | 1 | 16.5 | 39.5 | 147.8 | |
| 0.152 | 0.15 | 5.95 | 16.1 | 37 | 176.4 | |
| 0.0585 | 0.029 | 5.95 | 16.8 | 41.5 | 157.5 | |
| 0.152 | 0.075 | 1 | 16.8 | 41.5 | 276.3 | |
| 0.094 | 0.094 | 1 | 16.8 | 41.5 | 253.6 | |
| 0.0585 | 0.029 | 5.95 | 16.5 | 39.5 | 121.5 | |
| 0.094 | 0.094 | 6 | 15.7 | 34 | 91.5 | |
| 0.152 | 0.15 | 5.95 | 15.7 | 34 | 122.3 | |
| 0.094 | 0.047 | 1 | 16.8 | 41.5 | 196.8 | |
| 0.152 | 0.075 | 5.95 | 16.8 | 41.5 | 285.3 | |
| 0.094 | 0.094 | 6 | 16.5 | 39.5 | 185.6 | |
| 0.0585 | 0.058 | 5.95 | 16.8 | 41.5 | 184.9 | |
| 0.094 | 0.094 | 1 | 15.7 | 34 | 90.5 | D |
| 0.152 | 0.15 | 1 | 15.7 | 34 | 124.4 | |
| 0.152 | 0.15 | 1 | 16.8 | 41.5 | 361.5 | |
| 0.0585 | 0.058 | 5.95 | 15.7 | 34 | 70.91 | |
| 0.152 | 0.075 | 5.95 | 17.1 | 42.5 | 335.3 | |
| 0.152 | 0.15 | 1 | 16.1 | 37 | 182.4 | |
| 0.094 | 0.094 | 1 | 16.1 | 37 | 131.5 | |
| 0.094 | 0.094 | 6 | 16.8 | 41.5 | 244.6 | |
| 0.0585 | 0.029 | 5.95 | 17.1 | 42.5 | 180.5 | |
| 0.094 | 0.047 | 6 | 17.1 | 42.5 | 235.6 | E |
| 0.152 | 0.15 | 5.95 | 16.5 | 39.5 | 254.5 | |
| 0.094 | 0.094 | 1 | 16.5 | 39.5 | 191.6 | |
| 0.152 | 0.075 | 1 | 16.5 | 39.5 | 201.2 | |
| 0.0585 | 0.058 | 5.95 | 16.1 | 37 | 98.93 | |
| 0.094 | 0.094 | 6 | 16.1 | 37 | 127.5 | |
| 0.152 | 0.075 | 5.95 | 16.1 | 37 | 143.3 | F |
| 0.094 | 0.047 | 1 | 17.1 | 42.5 | 228.8 | |
| 0.152 | 0.075 | 1 | 17.1 | 42.5 | 325.3 | |

In the first experiment the data from **Table 3.** was partitioned into five sets described in **Table 4**. Each of five sets was divided into two subsets:

- *Training subset.* The 80% of the data in this subset is used to train the network.
  The rest of the data from the subset is used by the algorithm to compute the validation error. After each epoch of learning the learning algorithm uses the network to predict baring capacities using the validation data and computes the mean square error of predictions. If the error increases in 10 consecutives epochs, then the training is stopped (this technique allows to minimize the overfitting effect).
- *Test subset.* When the neural network is ready it is used to predict the values of bearing capacities given in the test subset.

**Table 4.** The data used for training MLPs in the first experiment.

| set number | training (80% training, 20% validation) | test |
|---|---|---|
| 1 | A+B+C+D+E | F |
| 2 | A+B+C+D | E+F |
| 3 | A+B+C | D+E+F |
| 4 | A+B | C+D+E+F |
| 5 | A | B+C+D+E+F |

In the second experiment only 6 samples from **Table 3.** are used to train DNNs. These samples are presented in **Table 5.** Five samples are used as training data and one sample is used as validation data.

**Table 5.** The data used for training DNNs in the second experiment

| B (m) | D (m) | L/B (-) | γ (kN/m³) | Φ (°) | $q_u$ (kPa) |
|---|---|---|---|---|---|
| 0.0585 | 0.029 | 5.95 | 15.7 | 34 | 58.5 |
| 0.0585 | 0.058 | 5.95 | 17.1 | 42.5 | 211 |
| 0.094 | 0.094 | 6 | 17.1 | 42.5 | 279.6 |
| 0.152 | 0.075 | 5.95 | 15.7 | 34 | 98.2 |
| 0.152 | 0.15 | 1 | 16.5 | 39.5 | 264.5 |
| 0.094 | 0.047 | 1 | 15.7 | 34 | 67.7 |

## 5. Results and discussion

The accuracy of predictions was measured by the absolute percentage error:

$$E_a = \frac{1}{n} \frac{|x_g - x_{nn}|}{x_g} \times 100,$$

where $x_g$ denotes the value of ultimate bearing capacity given in **Table 3.**, $x_{nn}$ is the value of bearing capacity predicted by the neural network and $n$ is number of samples in the test subset. Also, the maximal error was used:

$$E_{max} = \max_i \left\{ \frac{|x_g^{(i)} - x_{nn}^{(i)}|}{x_g^{(i)}} \times 100 \right\},$$

where $i$ goes over all samples in the test subset.

The accuracy of prediction was expressed as:

$$100\% - E_a$$

The results of the first experiment are given in **Table 6.** and **Figure 2.** There is no noticeable difference in the accuracy of deep and shallow networks trained on datasets 1.-4. As one would expect there is a decrease in the accuracy of neural networks related to the decrease in the number of training samples.

**Table 6.** Results of the first experiment

| | | set 1 | set 2 | set 3 | set 4 | set 5 |
|---|---|---|---|---|---|---|
| | | absolute error $E_a$ [%] (max error $E_{max}$ [%]) | | | | |
| number of hidden layers | 1 | 0.96 (1.47) | 1.17 (3.89) | 4.12 (12.96) | 5.64 (15.40) | 13.25 (55.31) |
| | 2 | 0.69 (2.01) | 1.10 (2.58) | 3.49 (10.13) | 4.13 (12.30) | 9.42 (10.78) |
| | 3 | 0.59 (0.83) | 1.34 (3.68) | 2.98 (8.59) | 4.27 (10.88) | 9.84 (11.53) |
| | 4 | 0.66 (0.92) | 0.70 (1.60) | 2.98 (9.03) | 3.80 (11.12) | 7.72 (17.84) |
| | 5 | 0.36 (0.62) | 0.79 (1.52) | 3.50 (8.63) | 4.20 (10.46) | 7.26 (18.72) |
| | 6 | 0.63 (1.05) | 0.79 (1.40) | 3.64 (10.15) | 3.81 (13.47) | 7.09 (15.72) |
| | 7 | 0.72 (0.89) | 0.67 (2.40) | 3.48 (12.11) | 4.17 (12.26) | 6.45 (13.78) |

On the other hand, there is a significant difference in the accuracy for the dataset 5. Shallow MLP (the best of 2000 trained shallow networks) gave the predictions with mean error of 13.25% (maximal error: 55.31%), while a deep neural network with 7 layers generated the predictions with mean absolute error of 6.45% (maximal error: 13.78%).

**Figure 2.** Chart representing results of the first experiment

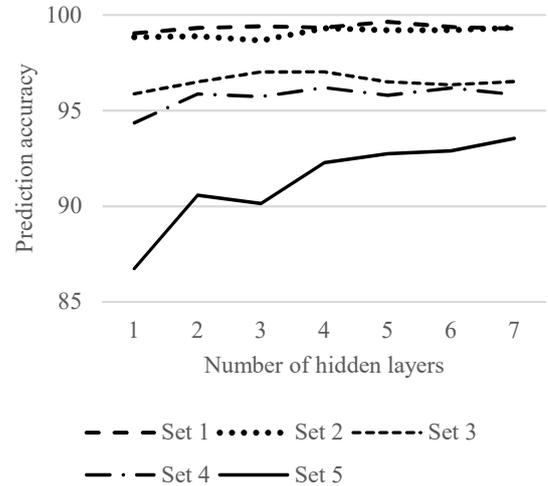

To investigate further the differences between shallow and deep neural networks trained on small datasets the second experiment was conducted. MLPs were trained on the dataset consisting of only 6 samples (**Table 5.**). The results of this experiment are presented in Table 7. The most accurate DNNs consisted of 5 to 7 layers. The minimal mean error (7.23%) was achieved by a DNN with 120 neurons collected in 5 layers. It is important to point out that MLPs with small (1,2) and large (10,15) number of layers performed significantly worse than DNNs with 5-7 layers.

**Table 7.** Results of the second experiment.

| | number of neurons | 90 | 120 | 150 |
|---|---|---|---|---|
| | | absolute error $E_a$ [%] (max error $E_{max}$ [%]) | | |
| number of layers | 1 | 21.63 (91.90) | 36.31 (115.7) | 55.67 (216.1) |
| | 2 | 11.37 (44.99) | 10.95 (42.11) | 15.32 (74.69) |
| | 3 | 9.51 (36.32) | 8.42 (29.46) | 11.30 (44.61) |
| | 4 | 9.44 (32.13) | 8.45 (27.58) | 8.85 (31.01) |
| | 5 | 9.61 (32.24) | 7.23 (26.70) | 8.38 (27.86) |
| | 6 | 9.98 (28.44) | 8.66 (24.55) | 9.22 (28.42) |
| | 7 | 8.97 (29.25) | 9.34 (26.18) | 8.18 (21.07) |
| | 8 | 9.13 (27.93) | 8.89 (27.68) | 8.62 (25.01) |
| | 9 | 9.84 (28.76) | 8.64 (26.81) | 8.54 (25.55) |
| | 10 | 10.51 (30.28) | 9.18 (27.64) | 9.18 (29.80) |
| | 15 | 21.73 (60.99) | 19.41 (68.87) | 21.23 (127.07) |

The predictions obtained by the best neural network from the second experiment are collected in **Table 8**.

**Table 8.** Results obtained by the most accurate neural network obtained in the second experiment.

| $q_u$ | DNN's prediction | $E_a$ [%] |
|---|---|---|
| 155.8 | 150.95 | 3.11 |
| 295.6 | 260.38 | 11.92 |
| 67.7 | 67.19 | 0.75 |
| 279.6 | 267.80 | 4.22 |
| 98.93 | 97.04 | 1.91 |
| 127.5 | 136.27 | 6.88 |
| 143.3 | 152.31 | 6.29 |
| 228.8 | 220.96 | 3.42 |
| 325.3 | 274.85 | 15.51 |
| 180.5 | 191.47 | 6.08 |
| 235.6 | 229.53 | 2.58 |
| 254.5 | 272.45 | 7.05 |
| 191.6 | 183.65 | 4.15 |
| 201.2 | 201.94 | 0.37 |
| 121.5 | 119.26 | 1.84 |
| 91.5 | 90.54 | 1.04 |
| 122.3 | 154.96 | 26.70 |
| 196.8 | 186.91 | 5.02 |
| 285.3 | 254.30 | 10.86 |
| 185.6 | 192.42 | 3.68 |
| 184.9 | 183.83 | 0.58 |
| 90.5 | 85.65 | 5.36 |
| 124.4 | 147.04 | 18.20 |
| 361.5 | 297.88 | 17.60 |
| 70.91 | 66.60 | 6.07 |
| 335.3 | 281.61 | 16.01 |
| 182.4 | 210.73 | 15.53 |
| 131.5 | 128.92 | 1.96 |
| 244.6 | 237.86 | 2.75 |
| 82.5 | 82.94 | 0.54 |
| 74.7 | 70.80 | 5.23 |
| 342.5 | 303.46 | 11.40 |
| 206.8 | 195.70 | 5.37 |
| 135.2 | 144.45 | 6.84 |
| 147.8 | 143.04 | 3.22 |
| 176.4 | 219.40 | 24.38 |
| 157.5 | 158.24 | 0.47 |
| 276.3 | 246.40 | 10.82 |
| 253.6 | 229.45 | 9.52 |
| 211.2 | 210.67 | 0.25 |
| 142.9 | 140.49 | 1.69 |
| 423.6 | 316.17 | 25.36 |
| 91.2 | 95.66 | 4.90 |
| 104.8 | 104.27 | 0.51 |

## 6. Conclusion

In this study the performance of deep and shallow MLPs trained on small data samples was tested. In more detail the prediction accuracy of deep and shallow MLP's was measured with the decrease of the training-to-testing data ratio. The following conclusions may be drawn based on the obtained results:

- The DNNs predictions based on small amount of training data are more accurate than the ones obtained by the shallow MLPs. This is a remarkable observation especially when modeling large scale or untypical physical systems. In those cases, experimental data is difficult to collect due to the economic aspects. Therefore, in such cases DNNs seem to a better choice for predicting bearing capacity.

- The ANNs may not require large amounts of training data to achieve high level of accuracy. In this study the accuracy of prediction seems to be more dependent on the number of layers in neural network than on the number of the neurons themselves. The research also shows that the networks' architecture may be of vital importance for the MLPs' performance in predicting bearing capacity. This MLPs' property is exposed in the part of the experiment with the low training-to-testing data ratio.

- As the study shows the best results are obtained by neural networks with 5 to 7 layers. Neural networks with less and – surprisingly – more layers obtain worse accuracy. It suggests that a 5,6,7-layered DNNs may be optimal for problems concerning predicting bearing capacity of shallow foundations.

- Given the accuracy of the DNN's it seems that obtaining high-quality learning datasets is more important than obtaining large learning datasets affected by a higher error. For example, it is worth considering conducting few experiments in the large scale than many laboratory-scale experiments as it is difficult to take into account all the factors influencing the prediction with the change of scale.